# Memory-augmented Chinese-Uyghur Neural Machine Translation

Shiyue Zhang[*], Gulnigar Mahmut[†], Dong Wang[‡], Askar Hamdulla[†]

[*]Beijing University of Posts and Telecommunications, Beijing, China
E-mail: byryuer@gmail.com
[†]Xinjiang University, Xinjiang, China
E-mail: {gulnigarmahmut, askarhamdulla}@sina.com
[‡]Tsinghua University, Beijing, China
E-mail: wangdong99@mails.tsinghua.edu.cn

*Abstract*— Neural machine translation (NMT) has achieved notable performance recently. However, this approach has not been widely applied to the translation task between Chinese and Uyghur, partly due to the limited parallel data resource and the large proportion of rare words caused by the agglutinative nature of Uyghur. In this paper, we collect ~200,000 sentence pairs and show that with this middle-scale database, an attention-based NMT can perform very well on Chinese-Uyghur/Uyghur-Chinese translation. To tackle rare words, we propose a novel memory structure to assist the NMT inference. Our experiments demonstrated that the memory-augmented NMT (M-NMT) outperforms both the vanilla NMT and the phrase-based statistical machine translation (SMT). Interestingly, the memory structure provides an elegant way for dealing with words that are out of vocabulary.

## I. INTRODUCTION

Uyghur belongs to the Turkic language, Altai family, and is primarily spoken by Uyghur people, an ethnic group in China with a population of more than 23 million, mostly distributed in the Xinjiang Uyghur Autonomous Region of China. Since people in most of the areas of China use Chinese, translation between Chinese and Uyghur becomes more and more important with the increase of cross-region collaboration and nation integration.

However, this translation turns out to be very challenging. One reason is that the two languages are rather "distant" in terms of both phonetics and linguistics. Particularly, the two languages have different syntactic orders: the order of Chinese is SOV (subject-object-verb), while the order of Uyghur is SVO. It implies that a phrase in Chinese probably cannot be translated into a single phrase in Uyghur, but some nonconsecutive words instead. As shown in Fig.1, a Chinese phrase "我爱" (I love) should be translated to two Uyghur words "مەن" (I) and "سۆيىمەن" (love), but these two words are far from each other in the corresponding Uyghur sentence. Unfortunately, this phenomenon is not infrequent, and it greatly increases the difficulty of the conventional phrase-based SMT that heavily relies on phrase correspondence thus weak in handling distant word reordering. Another difficulty of Chinese/Uyghur translation resides in the agglutinative nature of Uyghur, which means that Uyghur words can be produced in a very flexible way, by applying some general morphological rules. Put it more specific, Uyghur has about 30000 root words, 8 prefixes, more than 100 suffixes. A

Fig. 1 SOV in Chinese vs. SVO in Uyghur

Uyghur root can be agglutinated with unlimited number of suffixes, leading to a very large (even unlimited) vocabulary. For example, the Uyghur word "قىلالمىغانلىقىمدىنمىكىن" (Probably because I can't do it) has eight suffixes (ئال،مى،غان،لىق،ىم،دىن،مى،كىن) agglutinated to the root (قىل). A consequence of the flexible agglutination is that most of the Uyghur words are rare and even out-of-vocabulary (OOV) for an MT system.

Most of existing studies on Uyghur/Chinese translation focus on refining the phrase-based SMT architecture. For example, Dong et al. [1] tackled the syntactic order problem by reordering Chinese sentences to meet the SOV style of Uyghur, so that more correct phrase pairs could be obtained. Mi et al. [5] filtered out unreasonable phrase pairs from SMT phrase table by a binary classifier. Various morphology preprocessing techniques were also adopted to fit with Uyghur's agglutinative nature. These techniques segment Uyghur words into smaller morphemes: "stems", "suffixes" and "prefixes", and perform MT at the morpheme level [1, 2, 3]. These methods may produce better word alignment and alleviate the rare word and OOV word problem. All the above studies improved the Chinese/Uyghur translation to some extent, but none of them can be seen as a complete and systematic solution for the specific difficulties with these two languages.

Recently, neural machine translation (NMT) has attracted much attention. Different from SMT that relies on phrase pairs to perform token-by-token translation, NMT uses an encoder-decoder neural architecture to distill the meaning of the input, and then perform the translation by referring to the meaning discovered, leading to a meaning-oriented translation [16]. Particularly with the attention mechanism [6], the alignment between the source and target sentences can be very flexible. We conjecture this flexible alignment will provide a systematic solution for the syntactic order problem in the Chinese/Uyghur translation. For example, the Chinese word "我爱" (I love) can be aligned to two Uyghur words "مەن" (I) and "سۆيىمەن" (love) by the attention mechanism, no matter how distant the two words are in the Uyghur sentence.

However, applying NMT to Chinese/Uyghur translation is not as easy as the first glance. A major issue is that NMT has a tendency of overfitting to frequent words, while overlooking rare words that are not frequently observed. For Uyghur, unfortunately, most of the words are rare words. More seriously, the vocabulary of NMT systems cannot be too large due to the demand on computing resource. This leads to a severer OOV problem for Chinese/Uyghur translation, as many Uyghur words would become OOV with the limited vocabulary.

In this paper, we propose a memory-augmented NMT that equips the vanilla attention-based NMT with a memory component that stores some source-target word mappings. This design has two advantages: first, it inherits the superior of the attention-based NMT in dealing with the syntactic order problem; second, by using the memory, it can alleviate the rare word problem and provides an elegant treatment for OOV words. Our experiments showed that the M-NMT approach surpassed both the vanilla NMT and the SMT approaches.

## II. RELATED WORKS

This paper primarily follows our unpublished work in [7] where we demonstrated that a memory structure can significantly improve performance of Chinese-English translation. The idea of memory augmentation was inspired by the recent advances in Neural Turing Machine [8] and Memory Networks [9]. Our main difference from them is that the memory in the M-NMT model is fixed knowledge resevoir, rather than an updatable RAM. Arthur et al. [10] held an idea similar to ours and proposed to use some lexical knowledge to assist the translation. The difference is that we trained an independent attention on the lexical knowledge while they re-used the attention information of the NMT model.

Regarding OOV treatment, Gulcehre et al. [11] equipped NMT with a pointer network that can restore unknown words in the source context. This work was followed by Kong et al. [4], the only NMT study on Uyghur-Chinese translation we have found so far. Finally, Li et al. [12] proposed a replace-and-restore approach that replaces unknown words with similar words. Our model shares the same idea of using similar words to represent unknown words, but uses a pre-processing rather than a post-processing.

## III. ATTENTION-BASED NMT

Before introducing our M-NMT architecture, we will give a brief review of our reproduction of the attention-based NMT model [6]. This model has been regarded as being the state-of-art and will be used as a baseline in this study.

The attention-based NMT model is basically an encoder-decoder architecture, where the input word sequence $[x_1, x_2, ...]$ in the source language is embedded as a sequence of hidden states $[h_1, h_2, ...]$ by a bi-directional RNN with GRU as units, and another RNN is used to generate the target sequence $[y_1, y_2, ...]$. To guide the generation at each step to focus on a particular segment of source sentence, Bahdanau et al. introduced an attention mechanism. Specifically, when generating the i-th target word, the attention factor of the j-th source word is measured by the relevance between the current hidden state of the decoder, denoted by $s_{i-1}$ and the hidden state of the encoder at the j-th word $h_j$, given by:

$$\alpha_{ij} = \frac{e_{ij}}{\sum e_{ik}} \qquad e_{ij} = a(s_{i-1}, h_j)$$

where $a(\cdot,\cdot)$ is the MLP-based relevance function, and $\alpha_{ij}$ is the attention factor of $x_j$ at decoding step **i**. Then the semantic context that decoder focuses on is:

$$c_i = \sum \alpha_{ij} h_j$$

The decoder updates the hidden state with a recurrent function $f_d$, formulated by:

$$s_i = f_d(y_{i-1}, s_{i-1}, c_i)$$

and the next word $y_i$ is generated according to the following posterior:

$$p(y_i) = \sigma(y_i^T W z_i)$$
$$z_i = g(y_{i-1}, s_{i-1}, c_i)$$

where $\sigma(\cdot)$ is the softmax function, $g$ is a single maxout hidden layer. And being slightly different from the original model in Ref. [6] which takes $W$ as a parameter matrix, in our implementation, we take $W$ the same as the target word embedding $E_t$. Without performance loss, our modification decreases the number of parameters.

## IV. MEMORY-AUGMENTED NMT

### A. Architecture

The architecture of M-NMT is illustrated in Fig. 2. It involves two parts: the left part is a typical attention-based NMT which has been presented in Section III; the right part is a memory component. The outputs from the two parts are combined to produce the final translation.

As shown in Fig. 2, Memory has a hierarchical generation process. First of all, there is a big *global memory* (the bottom-right in Fig. 2) memorizing many source-target word mappings, written by $[y_{il}, x_i]$. One source word may have several possible translations. The *global memory* can be obtained from either a human-defined dictionary or the word table got from SMT. Human-defined dictionary is apparently better, because a word table from SMT may contains unreasonable mappings. However, collecting human-defined dictionary is expensive, and therefore we use word mappings produced by SMT in this study.

Next, based on the source words in each input sentence, appropriate elements in the *global memory* are selected dynamically to produce the *local memory*, as shown in the right-middle in Fig. 2. In order to involve the context information, $x_i$ is replaced by its hidden state $h_i$, written by $[y_{il}, h_i]$. Due to the limited computing resource, it is not possible to choose all the potential translations of each source word. We therefore select a few candidates according to $p(y_j|x_i)$, i.e., selecting the most possible translations. This selection method can be seen as a filter of the word mappings, with the aim of removing unreliable mappings obtained from SMT.

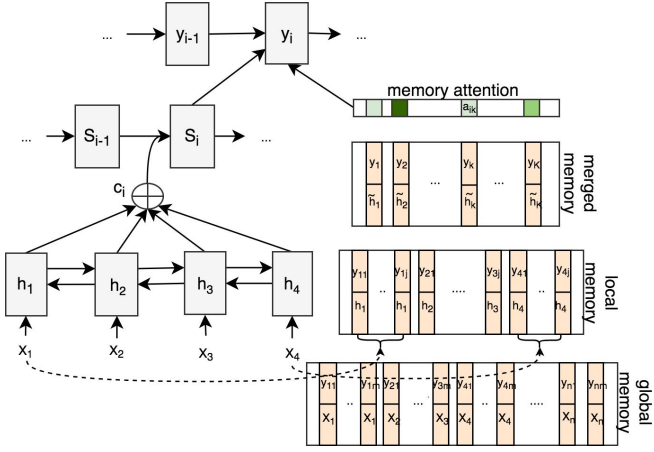

Fig. 2 The architecture of M-NMT

Finally, a particular target word may correspond to multiple source words in *local memory*. To save memory and load target words as many as possible, we propose a *merge* operation that consolidate elements with the same target word into a single element, given as follows:

$$u_k = \begin{bmatrix} y_j \\ \tilde{h}_i \end{bmatrix} = \begin{bmatrix} y_j \\ \sum_i p(x_i|y_j) h_i \end{bmatrix}$$

Noted that $p(y_j|x_i)$ and $p(x_i|y_j)$ are both obtained from word mappings in *global memory*. This leads to a *merged memory* as shown in Fig.2

### B. Memory Attention

In order to use the memory to improve translation, an attention mechanism is designed, following the same inspiration in the attention-based NMT. At each translation step **i**, denote the attention factor of each memory element $u_k$ by $\alpha_{ik}^m$, and assume it is derived from a relevance function $e_{ik}^m$:

$$\alpha_{ik}^m = \frac{e_{ik}^m}{\sum_{i=k}^{K} e_{ik}^m}$$

where K is the number of target words in the *merged memory*. In this study, $e_{ik}^m$ is designed as a simple neural net as follows:

$$e_{ik}^m = (v^m)^T \tanh(W_s^m s_{i-1} + W_u^m u_k + W_y^m y_{i-1})$$

where $s_{i-1}$ is the current decoder state, $y_{i-1}$ is the word generated at the previous step, $\theta^m = \{v^m, W_s^m, W_u^m, W_y^m\}$ represents the model parameters.

The attention factor $\alpha_{ik}^m$ can be used in different ways. Here we simply treat it as a posterior of the words in the vocabulary and combine it with the posterior $p(y_i)$ produced by the neural model part, resulting in a consolidated posterior, given by:

$$\tilde{p}(y_i) = \beta \alpha_{ik}^m + (1-\beta) p(y_i)$$

where $\beta$ is a pre-defined interpolation factor. This simple combination enables the memory component to be trained independently. The objective of the training is to let the memory attention as accurate as possible. Given the n-th training sentence, at each step **i**, the target attention should be 1 on the current word $y_i^n$ and 0 elsewhere. So, the objective function is defined as the cross entropy between the target attention and the memory attention, written by:

$$L(\theta^m) = \sum_n \sum_i \log\left(\alpha_{ik_i^n}^m\right)$$

where $k_i^n$ is the position of $y_i^n$ in the *merged memory*.

Note that joint training of the memory part and neural model part is possible, but it requires much more memory and suffers from the risk of overfitting. Therefore, in this study, we first train the neural model, and then keep it fixed and train the memory attention.

### C. OOV Treatment

The memory approach also provides an elegant way to address OOV words. In our study, OOV words are words absent in the vocabulary of the source or the target languages. Specifically, if an OOV word is encountered in the source sentence, the vector of a similar word is borrowed so that the OOV can be involved in the encoding. In contrast, if an OOV word is encountered on the target side during the memory construction, the vector of a similar word is likewise borrowed to represent the OOV word in memory. By this approach, OOV words can be represented, encoded and decoded as frequent words. Note that the selected similar word should not have appeared in the sentence, otherwise confusion will be caused. The similar word selection can be either manually defined by human, or based on word vector similarity. In this study, we simply use human definition.

## V. EXPERIMENTS

TABLE I
STATISTICAL INFORMATION OF CORPUS

| Uyghur | Sent. | Token. | Vocab. |
|---|---|---|---|
| Train | 180612 | 2868046 | 177420 |
| Dev | 1985 | 28945 | 10476 |
| Test | 992 | 15446 | 6666 |
| Chinese | Sent. | Token. | Vocab. |
| Train | 180612 | 2823872 | 130790 |
| Dev | 1985 | 28012 | 8706 |
| Test | 992 | 15009 | 5509 |

### A. Data

We collected more than 200,000 Uyghur-Chinese sentence pairs. After text cleaning, 183,589 sentence pairs were remained. We split the entire database into three sets: training (Train), development (Dev) and testing (Test). The statistical information of the three data sets is shown in Table I.

B. *Systems & Settings*

**SMT baseline**: For SMT, we used the state-of-art Moses toolkit [13]. We used SRILM [14] to train a 5-gram language model, and the GIZA++ toolkit to align the training data in both directions. The produced word alignments were used in both the SMT system and the M-NMT system.

**NMT baseline**: For NMT, we reproduced the attention-based NMT model [6]. The implementation was based on Tensorflow[1]. Both the encoder and encoder used one-layer RNN with GRU cells, and the encoder was a bi-directional RNN. The size of the vocabulary, the number of hidden units, and the number of output units were set to be 30000, 1000, 500 respectively. The batch size of the training was 80, and the optimization algorithm was Adam [15] with the learning rate setting to 0.0005. The decoding was based on a beam search algorithm, where the beam size was set to 12.

**M-NMT system**: The M-NMT system was the combination of the NMT baseline and a memory structure present in Section IV. When training the memory structure, i.e. the memory attention, the NMT component kept unchanged. The training and decoding settings were all the same as with the NMT baseline. The interpolation factor $\beta$ was set to $\frac{1}{3}$.

**Evaluation metrics**: For the evaluation, we use the BLEU score computed by *multi-bleu.perl* in *Moses*, which is the average of 1 to 4 gram BLEUs multiplied by a brevity penalty.

C. *Performance*

First of all, the whole training data, about 180,000 sentence pairs, was used to train both the Chinese-Uyghur and Uyghur-Chinese translation models, using SMT, NMT and M-NMT respectively. Table II presents the performance of these systems. The values shown in this table are BLEU scores.

TABLE II
PERFORMANCE OF DIFFERENT TRANSLATION SYSTEMS

| Systems | BLEU | | |
|---|---|---|---|
| | SMT | NMT | M-NMT |
| Ch-Uy | 32.44 | 35.24 | **36.88** |
| Uy-Ch | 32.24 | 36.67 | **37.86** |

As illustrated in Table II, when the whole training data was used, both NMT and M-NMT have a better performance than SMT. And our model, M-NMT, performs best. For Chinese-Uyghur translation, our model brings 4.41 and 1.66 BLEU improvement over SMT and NMT respectively. For Uyghur-Chinese translation, our model surpasses SMT and NMT by 5.62 and 1.19 BLEU score respectively.

To investigate why our model performs better, we observe the BLEU scores with different n-gram orders and the brevity penalty that trade-off the length of the translation. Table III shows results on the Chinese-Uyghur translation task. An interesting observation is that our model not only gives better 1-4 gram BLEU scores, but also has a more reasonable brevity penalty than NMT. This means that our model has alleviated the "under translation" problem of the vanilla NMT model.

TABLE III
1-4 GRAM BLEUS AND BREVITY PENALTY

| Systems | 1-gram BLEU | 2-gram BLEU | 3-gram BLEU | 4-gram BLEU | Brevity penalty |
|---|---|---|---|---|---|
| **SMT** | 54.5 | 34.6 | 26.6 | 22.1 | **1.000** |
| **NMT** | 57.7 | 39.8 | 31.9 | 27.0 | 0.939 |
| **M-NMT** | **58.8** | **40.8** | **32.4** | **27.1** | 0.968 |

To test our hypothesis that our model provides better treatment for rare words, we count the number of "recalled words" in the test set, e.g. the number of unique words appeared in the translation. In the Chinese-Uyghur translation task, the reference of the test set contains 6666 unique words. In the translation result of the SMT, NMT and M-NMT systems, the numbers of recalled words are 3680, 3509, 3560 respectively. It can be seen that our system recalls more words than the NMT system but less than the SMT system, which is consistent with our expectation that our model can alleviate the rare word problem of NMT.

Besides, we study the influence of the volume of the training data. As shown in Table IV, we compare the performance when 100%, 50% and 25% of the training data are used to build the three systems. It can be observed that NMT can perform better than SMT even if half of the training data, ~90,000 sentence pairs, are used. When the 25% of the training data are used, NMT cannot surpass SMT, but our M-NMT model still outperforms SMT. This means that M-NMT can leverage the power of SMT by using the memory structure (produced by SMT) and deliver good performance even with limited training data.

TABLE IV
PERFORMANCE WITH DIFFERENT AMOUNT OF TRAINING DATA

| Data used | BLEU | | |
|---|---|---|---|
| | SMT | NMT | M-NMT |
| 100% | 32.44 | 35.24 | **36.88** |
| 50% | 21.97 | 22.30 | **23.70** |
| 25% | 17.30 | 16.75 | **18.72** |

Finally, we give an example to demonstrate that the M-NMT model, accompanied by the similar word approach, can deal with OOV words. The source sentence is "黎族 的 婚礼 别具一格 。", which means "Lizu's wedding is unique". The word "黎族" (Lizu, an ethnic group in China) is an OOV word. We use a similar word in the vocabulary "撒拉族" (another ethnic group in China) to represent this OOV word. By this similar word replacement, "سالا", the translation of "撒拉族", is obtained but redirected to the correct word "لزۇ", the correct translation of the original OOV word "黎

---

[1] https://www.tensorflow.org/

族". The translation result is shown in Fig. 3. It can be seen that the M-NMT model obtains a correct translation.

Src:             黎族 的 婚礼 别具一格 。
Ref:             لىزۇ مىللىتىنىڭ توي مۇراسىمى تولىمۇ ئۆزگىچە .
NMT:             UNK_ توي مۇراسىمى ئۇتكۇزۇۋشىنىڭ ئالدىنقى شەرتى قىلىنىدۇ .
M-NMT+OOV:       لىزۇ مىللىتىنىڭ توي مۇراسىمى تولىمۇ ئۆزگىچە .

Fig. 3 An example of M-NMT OOV treatment

## VI. CONCLUSIONS

In this paper, we demonstrated that the attention-based NMT approach can perform well in translation tasks between Chinese and Uyghur, partly due to the attention mechanism that can solve the synaptic order discrepancy between the two languages. Additionally, we introduced a memory-augmented NMT, which equips NMT with an external memory to memorize some translation pairs. We conjecture that this memory may alleviate the rare word problem caused by the agglutinative nature of Uyghur. Our experiment demonstrated the M-NMT approach can significantly improve Chinese-Uyghur translation, especially on rare words. It also demonstrated that the memory mechanism provides an elegant way for OOV treatment when accompanied by a similar word strategy.

Two drawbacks exist in our model. Firstly, the quality of the OOV treatment heavily relies on the quality of the similar word selection. If the similar words are not truly similar, the performance will drop dramatically. A better OOV word embedding approach will be investigated to improve the OOV treatment. Secondly, the present memory structure does not include any phrase mapping, though the advantage of using phrases is well known in MT. We will extend word memory to phrase memory in the future work.

## REFERENCES

[1] Dong, X., Xue, H., Ma, B., & Wang, L. (2010). Chinese-uyghur statistical machine translation: The initial explorations. Universal Communication Symposium (pp.320-324). IEEE.
[2] Dong, X., Xue, H., & Yang, Y. (2012). Factor-based uyghur-chinese statistical machine translation. *International Journal of Advancements in Computing Technology, 4*(2), 275-283.
[3] Aisha, B., & Sun, M. (2010). Uyghur-chinese statistical machine translation by incorporating morphological information. *Journal of Computational Information Systems*.
[4] Kong, J., Yang, Y., Zhou, X., Wang, L., & Li, X. (2016). Research for Uyghur-Chinese Neural Machine Translation. Natural Language Understanding and Intelligent Applications. Springer International Publishing.
[5] Mi, C., Yang, Y., Zhou, X., Wang, L., Li, X., & Tursun, E. (2014). A phrase table filtering model based on binary classification for uyghur-chinese machine translation. *Journal of Computers, 9*(12).
[6] Bahdanau, D., Cho, K., & Bengio, Y. (2014). Neural machine translation by jointly learning to align and translate. *Computer Science*.
[7] Yang F., Shiyue Z., Andi Z., Wang D. & Andrew A. Memory-augmented Neural Machine Translation. unpublished.
[8] Graves, A., Wayne, G., & Danihelka, I. (2014). Neural turing machines. *Computer Science*.
[9] Weston, J., Chopra, S., & Bordes, A. (2014). Memory networks. *Eprint Arxiv*.
[10] Arthur, P., Neubig, G., & Nakamura, S. (2016). Incorporating discrete translation lexicons into neural machine translation.
[11] Gulcehre, C., Ahn, S., Nallapati, R., Zhou, B., & Bengio, Y. (2016). Pointing the Unknown Words. *Meeting of the Association for Computational Linguistics* (pp.140-149).
[12] Li, X., Zhang, J., & Zong, C. (2016). Towards zero unknown word in neural machine translation. *International Joint Conference on Artificial Intelligence* (pp.2852-2858). AAAI Press.
[13] Koehn, Philipp, Hoang, Hieu, Alexandra, & CallisonBurch, et al. (2007). Moses: open source toolkit for statistical machine translation. *in Proceedings of the Association for Computational Linguistics (ACL'07,9*(1), 177--180.
[14] Stolcke, A. (2002). Srilm --- An Extensible Language Modeling Toolkit. *International Conference on Spoken Language Processing* (pp.901--904).
[15] Kingma, D. P., & Ba, J. (2014). Adam: a method for stochastic optimization. *Computer Science*.
[16] Ilya Sutskever, Oriol Vinyals, and Quoc V Le. (2014). Sequence to sequence learning with neural networks. In Advances in neural information processing systems, pp. 3104–3112.